\documentclass[final]{cvpr}

\usepackage{times}
\usepackage{epsfig}
\usepackage{graphicx}
\usepackage{amsmath}
\usepackage{amssymb}
\usepackage{multirow}
\usepackage{array}

\usepackage[pagebackref=true,breaklinks=true,colorlinks,bookmarks=false]{hyperref}

\def\cvprPaperID{5180} 
\def\confYear{CVPR 2021}

\begin{document}

\title{Inverting Generative Adversarial Renderer for Face Reconstruction}
\author{
Jingtan Piao\textsuperscript{1},
~~~  Keqiang Sun\textsuperscript{1},
~~~ Quan Wang\textsuperscript{2,3},
~~~ Kwan-Yee Lin\textsuperscript{1,2}\thanks{K. Lin and H. Li are the co-corresponding authors.},
~~~ Hongsheng Li\textsuperscript{1,4}\footnotemark[1] \\
\textsuperscript{1}CUHK-SenseTime Joint Laboratory, The Chinese University of Hong Kong\\
\textsuperscript{2}SenseTime Research and Tetras.AI~~~
\textsuperscript{3}Shanghai AI Laboratory~~~
\textsuperscript{4}School of CST, Xidian University \\ 
{\tt\small \{1155116308, kqsun\}@link.cuhk.edu.hk,  \{wangquan, linjunyi\}@sensetime.com, hsli@ee.cuhk.edu.hk}
}

\maketitle
\pagestyle{empty}
\thispagestyle{empty}

\begin{abstract}

Given a monocular face image as input, 3D face geometry reconstruction aims to recover a corresponding 3D face mesh.
Recently, both optimization-based and learning-based face reconstruction methods have taken advantage of the emerging differentiable renderer and shown promising results.
However, the differentiable renderer, mainly based on graphics rules, simplifies the realistic mechanism of the illumination, reflection, \etc, of the real world, thus cannot produce realistic images.
This brings a lot of domain-shift noise to the optimization or training process.
In this work, we introduce a novel Generative Adversarial Renderer (GAR) and propose to tailor its inverted version to the general fitting pipeline, to tackle the above problem.
Specifically, the carefully designed neural renderer takes a face normal map and a latent code representing other factors as inputs and renders a realistic face image.
Since the GAR learns to model the complicated real-world image, instead of relying on the simplified graphics rules, it is capable of producing realistic images, which essentially inhibits the domain-shift noise in training and optimization.
Equipped with the elaborated GAR, we further proposed a novel approach to predict 3D face parameters, in which we first obtain fine initial parameters via Renderer Inverting and then refine it with gradient-based optimizers.
Extensive experiments have been conducted to demonstrate the effectiveness of the proposed generative adversarial renderer and the novel optimization-based face reconstruction framework. Our method achieves state-of-the-art performances on multiple face reconstruction datasets.
\vspace{-0.3cm}
\end{abstract}
\vspace{-0.3cm}
\section{Introduction}

Faithfully recovering the 3D shapes of human faces from unconstrained 2D images is a challenging task and has numerous applications such as face recognition and face animation~\cite{kaisiyuan2020mead,aot2020neurips}.
State-of-the-art 3D face reconstruction methods can be generally categorized into two groups, learning-based methods and optimization-based methods.

\begin{figure}[th]
\begin{center}
	\includegraphics[width=1.0\linewidth]{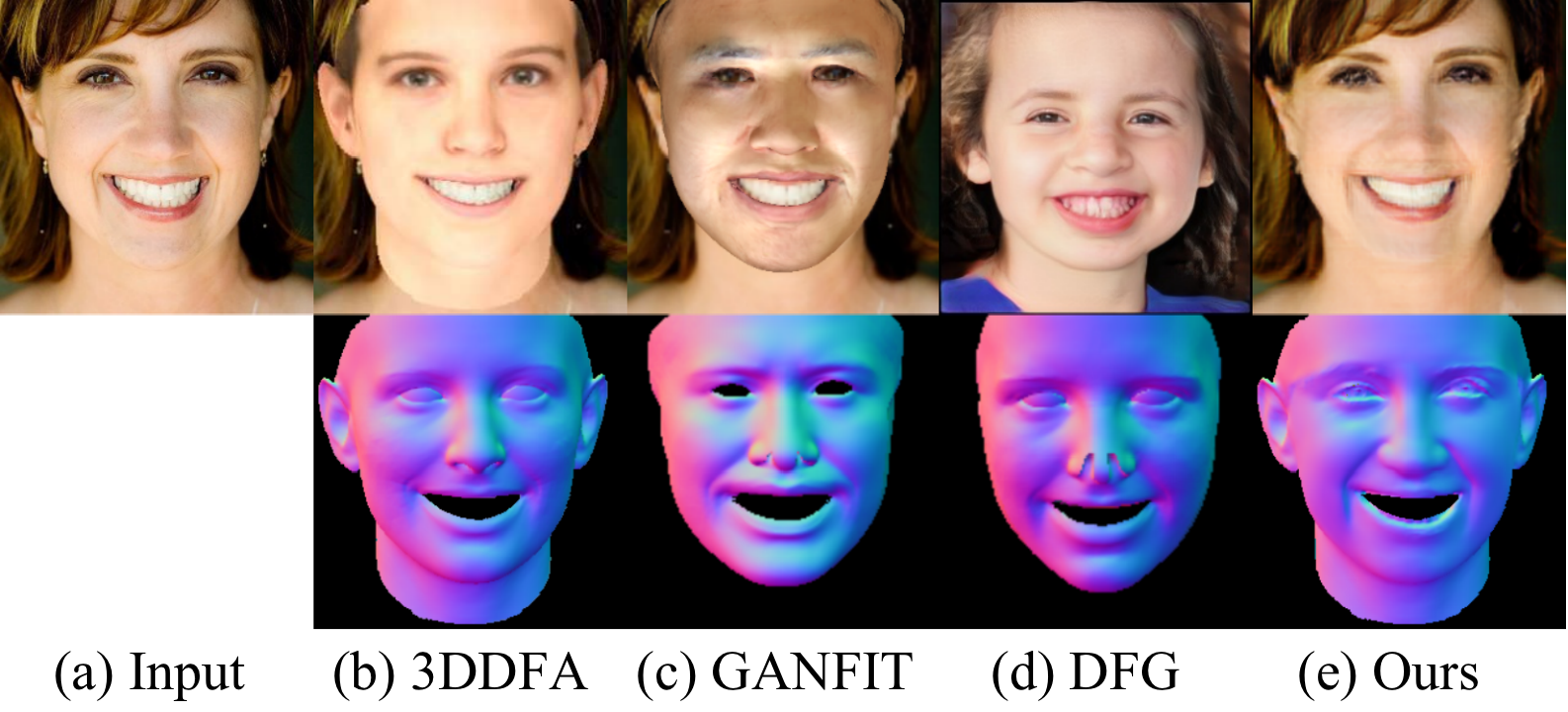} \hspace{10pt}
\end{center}
\vspace{-0.7cm}
\caption{Comparisons with state-of-the-art face renderers. On the second row are input geometry and the first row are corresponding rendered images. Output of (b)~\cite{zhu2016face} and (c)~\cite{gecer2019ganfit} are not realistic, since they use graphics-based renderers. And there may exists inconsistency between the input and the rendered image in (d)~\cite{deng2020disentangled}. Our method faithfully renders realistic images consistent with the input geometry, as shown in (e).}
\label{fig:motivation}
\vspace{-0.4cm}
\end{figure}

The deep learning-based methods~\cite{zhu2016face,cao2014displaced,genova2018unsupervised,deng2019accurate} usually take place in a regression manner, which takes facial images as inputs and learn to regress the corresponding 3DMM parameters. However, these methods usually require large amounts of labeled data, while the ground truth 3DMM parameters are rather difficult to acquire.
Optimization-based methods~\cite{blanz2003reanimating,kato2018neural,gecer2019ganfit,zhu2020reda}, on the other hand, generally treat the imaging of faces as a generative process \cite{mirza2014conditional}, which takes a series of geometry coefficients (\eg, albedo, texture, lighting, viewing angle, etc.) as inputs and outputs a rendered image according to certain graphics rules. The distances between the rendered images and the target images are minimized with an optimization framework.
However, since the graphics rules generally employ simplified models to characterize the physical process of capturing face images, many details of the imaging process cannot be modeled, which introduces difficulties for the optimization of face reconstruction.

Recent developments of the differentiable renderers provide an efficient tool for both types of face reconstruction methods.
Specifically, the regressed parameters in learning-based methods could be rendered to images, with which the photometric loss can be adopted for optimization. In this manner, as shown in~\cite{deng2019accurate}, learning-based models may be trained without geometry ground truth of the input image.
For the optimization-based methods, as introduced by~\cite{gecer2019ganfit}, differentiable renderers introduce gradient-based optimization and allow adopting more complicated losses and stabilizes the training process.

However, differentiable renderers have two drawbacks. On the one hand, the differentiable renderers are created by handcrafted rendering rules and are generally not capable of producing realistic images. The domain gap between the rendered and real images hinders the optimization or the training process. On the other hand, the differentiable renderers are difficult to optimize as they can only back-propagate errors to local vertices. As shown in (b) and (c) of Figure~\ref{fig:motivation}, the rendered image is not realistic since they are using graphics-based renderers.
Some methods \cite{kato2018neural, liu2019soft} modify the renderers to make them ``more'' differentiable and better converge to the optimum via optimization, whereas they are still utilizing the graphics-based rendering methods, hence the above two problems remain essential drawbacks of the differentiable renderer.


An intuitive solution is to replace the differentiable renderer with a neural renderer, an emerging method to employ a neural network to render an image corresponding with the given geometry and texture conditions. Actually, several types of neural renderers have been proposed and studied before.
For instance, 
Deng \etal \cite{deng2020disentangled} proposed a neural renderer, which takes 3DMM parameters as inputs and generates a facial image. Nevertheless, the 3DMM parameters are too abstract for the control of the generative adversarial renderer. Therefore, the rendered images, although are more realistic and basically subject to the inputs, do not strictly condition on the 3DMM parameters. As shown in (d) of Figure~\ref{fig:motivation}, even though the input geometry parameters are close to the target person, the rendered image shows a large variation. Hence, it is not an ideal neural renderer for face reconstruction.

In this paper, we propose to adopt a novel conditional neural renderer, trained in a self-supervised manner, to replace the conventional graphics-based differentiable renderer, to tackle the aforementioned problems while maintaining the advantages of utilizing a renderer for training.
The proposed conditional face neural renderer takes a face normal map as the geometry condition and a latent code vector to model other influencing factors. Since we hope the proposed renderer could facilitate the optimization of the face geometry, we decouple the normal map from the other condition factors so that the geometry could be better reconstructed via optimization of the normal map. 
To further enhance the controllability of the normal map upon the rendered images, a novel Normal Injection Module (NIM) is proposed, in which the normal map is used to modulate the convolution kernel by pixel-wise multiplication on each channel, to determine the geometry.
On the other hand, the decoupled latent code contains detailed information about the facial textures, which are also significant in reconstructing the image faithfully. 
With a novel normal consistency loss, the whole neural renderer is trained in a self-supervised manner without any labeled data.
As shown in (e) of Figure~\ref{fig:motivation}, the proposed GAR could faithfully render a realistic face image, according to the input geometry map.

After the neural renderer is trained, it takes the place of the differentiable renderer in the optimization-based face geometry reconstruction pipeline, in which the deviation between the given image and the rendered image is minimized and the geometry corresponding to the normal map is optimized.


Even with the proposed neural renderer, direct optimization with random initialization still struggles to recover the optimal 3D face shape.
We further proposed a novel approach to predict 3D face parameters, in which we first predict a set of good initial 3D parameters by a separate neural network and then refine them with a gradient-based optimizer.
Inspired by the latest GAN inverting technique \cite{bau2019seeing}, we train a regression network to predict a good initialization of the latent code for inverting the neural renderer to robustly recover the conditioning face normal map.
The optimal face normal maps and subsequently the corresponding face shapes can then be obtained via iterative gradient-based optimization.

\begin{figure*}[th]
\begin{center}
	\includegraphics[width=0.9\linewidth]{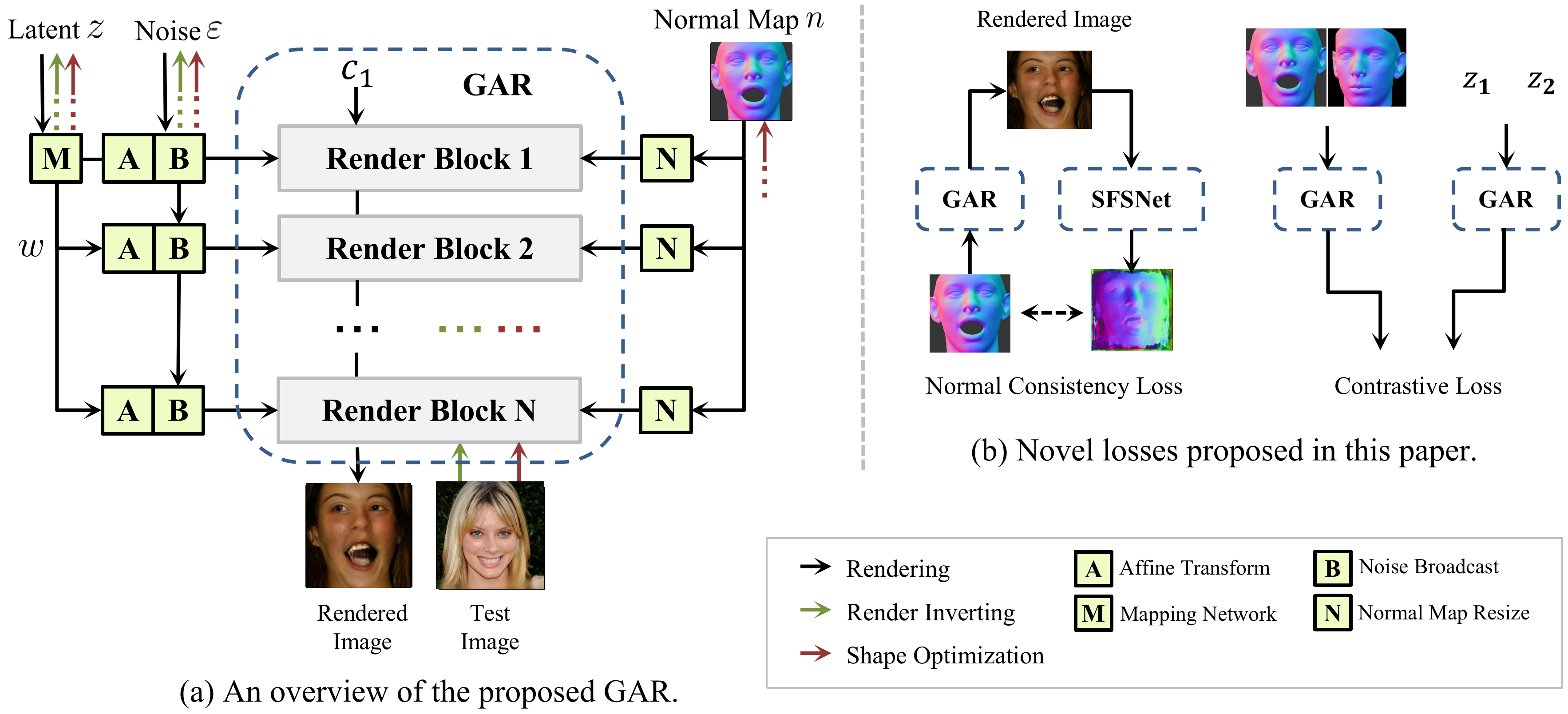} \hspace{10pt}
\end{center}
\caption{The architecture of the proposed method.
(a) The proposed Generative Adversarial Renderer (GAR) is composed of a series of Render Blocks. The latent code $z$, mainly encoding non-geometry information, is transformed to the normalization parameters $w$ through a mapping network $M$. The noise code $\varepsilon$ is broadcast to each block with $B$. And the normal map is resized to certain resolutions with $N$. (b) In training, we introduced a novel Normal Consistency Loss to enhance the controllability of the input normal map, and a Contrastive Loss to decouple the normal $n$ and the latent code $z$.}
\label{fig:stylegan}
\end{figure*}

The proposed optimization algorithm has two unique advantages.
1) The optimization process is more stable because the ``fully'' differentiable neural renderer has larger receptive fields and can achieve more accurate image reconstruction.
2) With the proposed initialization-prediction network, the neural renderer can be easier inverted to convergence and achieve better accuracy on face reconstruction.

In summary, the main contributions of the proposed method are three-fold:
\begin{itemize}
	\item To the best of our knowledge, we are the first to employ a conditional neural renderer, instead of a graphics-based differentiable renderer, to facilitate the face reconstruction.

	\item We propose a novel normal-conditioning neural renderer that can produce vivid face images conditioned on the input normal map and a latent code.
	\item We propose a face reconstruction algorithm based on the novel neural renderer, and achieve state-of-the-art performance on multiple face reconstruction datasets.
\end{itemize}


\section{Related Work}

\subsection{3D Face Modeling}

Face modeling aims at using mathematical formulas to generate locations of vertices of a face mesh. Since the introduction of the Basel Face Model \cite{blanz1999morphable} that used a linear combination of Gaussian distributed coefficients of a face, multiple methods have been proposed, including blender shapes \cite{cao2014facewarehouse}, skeleton animations \cite{li2017learning}, B-splines \cite{magnenat1988abstract}, deformation based method \cite{jiang2019disentangled}. Nonlinear models have also been proposed~\cite{tran2018nonlinear,tran2019towards}. However, all of the face models can only formulate the facial regions and lack the modeling of the face accessories and hairs.
Some efforts have been made to model various face attributes, including hair \cite{hu2015single}, mouth \cite{nagano2018pagan}, back of the head~\cite{ploumpis2020towards,lattas2020avatarme}, \etc.
However, they are generally more complicated and require much more computation to fit or render a face image.

\subsection{3D Face Reconstruction}
3D face reconstruction is the inverse process of recovering the face shapes from a monocular image. The methods can be generally categorized into two types, optimization-based methods and learning-based methods. The optimization-based methods generally try to invert the rendering or imaging process of face images by optimizing a cost function for each input image~\cite{wu2018alive,ploumpis2020towards}. Various loss functions were explored, including the re-projection loss of the detected 2D landmarks from the rendered 3D mesh \cite{blanz2003reanimating}, the photometric loss between the rendered image by a differentiable renderer and the original image \cite{kato2018neural}, etc.
\cite{gecer2019ganfit} introduces a pretrained generative adversarial network to fit the texture UV map. And thanks to the differential renderers, \cite{gecer2019ganfit} adopts photo-metric loss, as well as recognition loss to further enhance the texture and geometry quality.
\cite{zhu2020reda} proposes a ReDA Rasterizer for more soft and realistic rendering, and a free-form deformation layer with as-rigid-as-possible constraint to reconstruct an accurate face model.

Recently, deep learning-based methods have presented promising performance. Zhu \etal \cite{zhu2016face} proposed to directly regress the face parameters from the input face image. Starting from the cascaded face-parameter regressor \cite{cao2014displaced}, there are methods focusing on designing supervisions on representing the final reconstructed mesh. Video-based methods introduced additional constraints to regularize the reconstruction results. \cite{genova2018unsupervised} assumed that the reconstructed face images from multiple frames of a video should maintain the same face identity and similar textures. Face recognition models and perceptual loss are adopted to minimize the differences between the feature maps of the multi-view images to better regularize the reconstructed face shapes.


\subsection{Face Image Generation}
Many 3D-based methods have been proposed to generate face images~\cite{gecer2020synthesizing,deng2020disentangled}.
Besides methods using a statistic face model to explicitly calculate the 3D mesh, face image auto-encoder has been popular since the introduction of generative adversarial network \cite{goodfellow2014generative}. High-resolution images can be gradually generated from a low-resolution to high-resolution manner. StyleGANs \cite{karras2019style,Karras2019stylegan2} introduced to model the latent code for image generation as the input Batch Normalization parameters.
There are also methods that try to control the generated images with the input latent code \cite{mirza2014conditional} by adding a classifier to restrict the output patterns.
However, the controllable properties of the generated face images are only limited to the modification of single neurons \cite{shen2019interpreting}. Some feature disentanglement networks \cite{abdal2019image2stylegan} for image generation have also been proposed. However, they cannot generate realistic images conditioned on 3D information.
\cite{nguyen2019hologan} tried to disentangle the process of 2D face image generation into 3D mid-level feature generation and 3D-to-2D feature projection and generation.
Other than how to generate more realistic face images, given a trained GAN model, how to effectively invert the GAN to obtain the corresponding latent code is also of importance.
\cite{bau2019seeing} studied how to effectively invert a GAN model, since the inverting of the GAN model might also be stuck at local minima. A regression network is trained with the generator network predicting a good initialization for inverting the GAN from the input image to recover the corresponding latent code.

\section{Method}

The goal of this work is to reconstruct its corresponding face geometry parameters from a single image.

Given an image,
1) the latent code and noise will be initialized via GAR Inverting network, and 
2) 3DMM parameters will be initialized with the fitting method.
With parameters initialized from 1) and 2), we then finetune all parameters and latent codes by back-propagation.

The key of the proposed algorithm is a Generative Adversarial Renderer $G$, which is trained to generate realistic face images conditioned on the input face normal map and a latent code.
$G$ is trained in a self-supervised manner, and no labeled data is required.
The renderer $G$ is fixed after training. Given an unseen face image, a renderer inverting network $R$ is trained to predict a good initialization for the latent code, based on which, a gradient-based optimizer can effectively recover the face geometry parameters.

\subsection{Generative Adversarial Renderer}

In this section, we introduce the proposed Generative Adversarial Renderer $G$, which takes in a normal map $n$ and latent code $z$ and outputs a corresponding rendered image $I_{out}$.

\begin{figure}[t]
\begin{center}
	\includegraphics[width=0.9\linewidth]{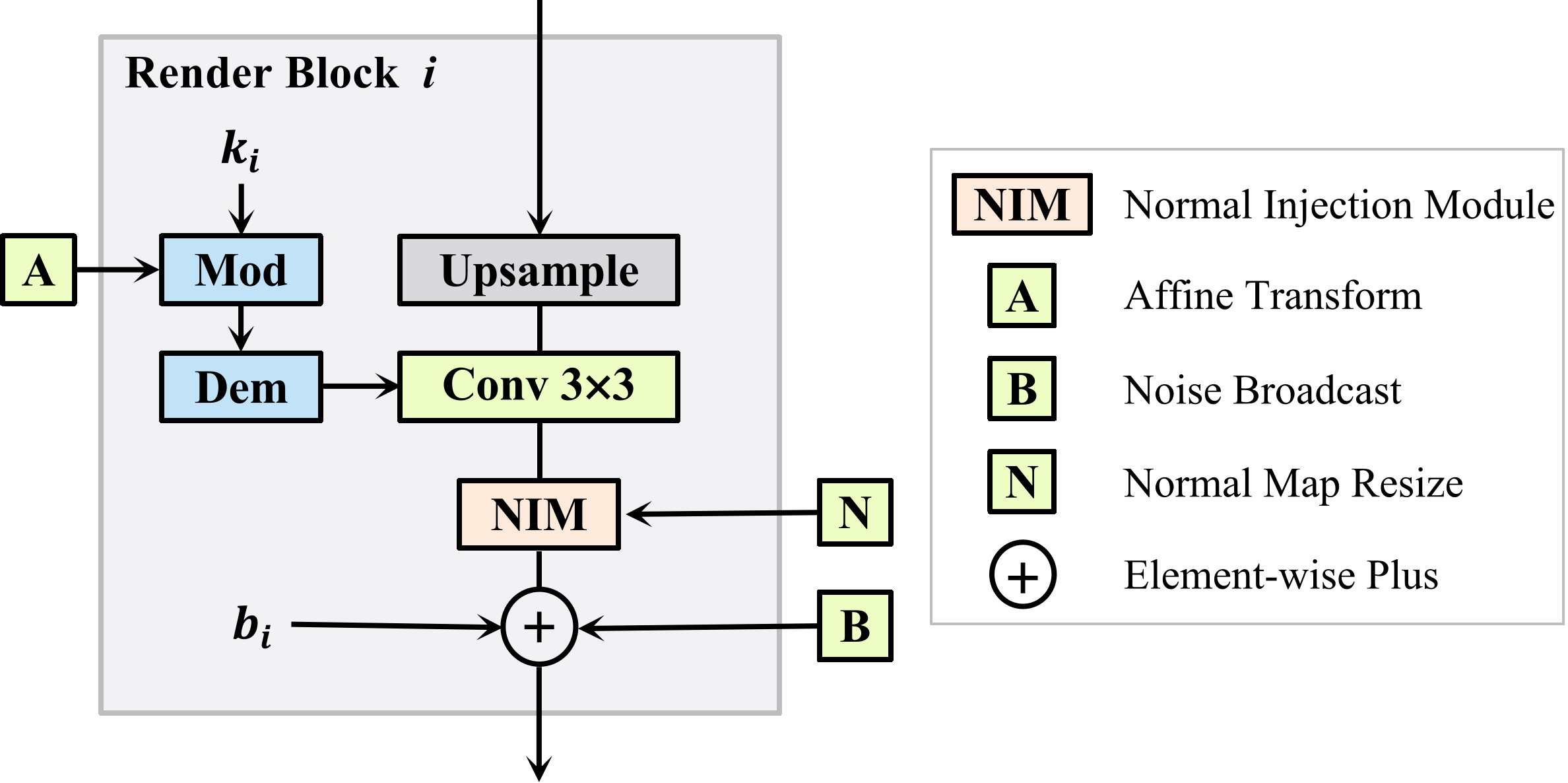} \hspace{10pt}
\end{center}
\caption{Details of the Render Block. The Render Block is based on the StyelGan v2 block~\cite{Karras2019stylegan2}, in which we introduce a Normal Injection Module to take in the resized normal map.}
\label{fig:render_block}
\end{figure}

\noindent
\textbf{Architecture.} The proposed Generative Adversarial Renderer $G$ is composed of a series of Render Blocks, based on StyleGan v2~\cite{Karras2019stylegan2}, as shown in Figure~\ref{fig:stylegan} (a). Each block, corresponding to a certain resolution, contains style-varying convolutions, modulated by a latent code $w$ mapped from an input latent code $z$. The feature map after the convolution is then modulated by a normal map $n$, which can be generated from 3DMM face models with different shape $\alpha$, expression $\beta$, and pose $\theta$ parameters. See Figure~\ref{fig:render_block} for details.


The latent code $z$, which encodes factors of a face image other than its normal, is transformed to the normalization parameters $w$ through a mapping network $M$ for modulating kernel parameters of convolution in each block of the network. The modulation and demodulation of the style-varying kernel parameters $k$ by the latent code $z$ is defined as
\begin{equation}
\begin{split}
	k'_{cij} = w_{c}k_{cij}/\sqrt{\sum_{c,j}(w_{c}k_{cij})^{2} + \epsilon},
\end{split}
\label{equ:1}
\end{equation}
\begin{equation}
\begin{split}
	w = M(z),
\end{split}
\label{equ:2}
\end{equation}
\noindent
where $k_{cij}$ denotes the initial kernel parameter at spatial position $(i,j)$ of the $c$-th channel, $k'_{cij}$ denotes the kernel parameter after modulation, $w_{c}$ is the modulation parameter for the $c$-th instance channel, predicted from the latent code $z$ by an 8-layer MLP $M$, as represented in Equation~\ref{equ:2}. And $\epsilon$ here is used to avoid numerical division by zero.

The input feature map $f_{cxy}$ is convoluted to $f'_{lxy}$ with the modulated kernels

\begin{equation}
\begin{split}
	f'_{lxy} = \sum_{i,j}{k^l_{cij}f_{c,x+i,y+j}},
\end{split}
\end{equation}

\noindent
where $f'_{lxy}$ indicates the feature map pixel at $(x, y)$ in the $l$-th channel.

The normal map is used to further modulate the feature map $f'$ in the Normal Injection Module (NIM).
However, instead of regularizing the channel dimension, the normal map is used for regularizing the spatial dimension:
\begin{align}
	f''_{lxy} = n_{xy}f'_{lxy},
\end{align}
where $f''_{lxy}$ is the feature maps after modulation by the input normal map, and $n_{xy}$ denotes the injecting normal values from the facial normal map $n$ from the spatial location $(x, y)$.

The feature maps $f''$ are added with a learned bias $b$ and a random Gaussian noise map $\varepsilon$ before sent to the next block. The insight here, as discussed in the original StyleGAN~\cite{karras2019style}, is that the small dimension of the latent code $z$ cannot fully express all the details of a face image. The extra noise is therefore needed to properly model the extra information.

\noindent
\textbf{Loss Functions.}
Given a normal map $n$, a latent code $z$ and a random noise $\epsilon$, the neural renderer outputs a corresponding face image,
\begin{align}
	&I_{\rm out}  = G(n, z, \varepsilon).
\end{align}
Besides the commonly used adversarial loss for encouraging image vividness, to regularize the results of $G$ to match the input conditioning normal map, we propose a cycle normal consistency loss (see the left side of Figure~\ref{fig:stylegan} (b)).
A pre-trained face normal estimation network $N$ is employed to predict the face normal map of the generated image $I_{\rm out}$.
Intuitively, if the generated image well fits the conditioning normal map, the face estimation network would estimate very similar face normal map to the input one. Thus we define the normal consistency loss as
\begin{align}
	\mathcal{L}_{n} = &\|P(I_{\rm out}) \odot (n - N(I_{\rm out}))\|_1,
\end{align}
where $I_{\rm out}$ is the rendered face image, $P(\cdot)$ is a face parsing network \cite{badrinarayanan2017segnet} that outputs the facial region mask and makes the loss only effective on the facial region, $\odot$ denotes element-wise multiplication, and $N(\cdot)$ denotes the pre-trained face normal estimation network.
We leverage SFSNet~\cite{sengupta2018sfsnet}, which is trained with synthetic images and unlabeled real images, as the normal estimation network here.



Since the proposed neural renderer targets the face geometry reconstruction, we would like the face shape to be neatly controlled by the input face normal map but not the latent code $z$. 
Therefore, we introduce two contrastive losses to facilitate the disentanglement and to strengthen the controllability of the input normal map (see the right side of Figure~\ref{fig:stylegan} (b)).

On the one hand, we would like the face structure to be fully controlled by the input normal map. We construct pairs training data, in which the paired data have identical normal maps $n_1 = n_2$ but their latent codes $z_1$ and $z_2$ are different. In addition, we further adopt a facial landmark detector as a measurement of the structure consistency. Facial landmarks are essential complements to the normal maps, since normal maps focus on general structures of the surface, while the landmarks pay more attention to the facial edges and boundaries. Specifically, the facial landmark consistency loss $\mathcal{L}_{ldmk}$ is formulated as
\begin{align}
	&\mathcal{L}_{\rm ldmk}( n, z_{1}, z_{2}) = \nonumber \\
	&\|L(G( n, z_{1}, {\varepsilon})) - L(G( n, z_{2}, {\varepsilon}))\|_2,
\end{align}
where $L$ is the pre-trained landmark detector, $z_{1}$, $z_{2}$ are different latent codes fed into the renderer $G$.

On the other hand, we require to retain the identity of the same person when his/her pose $\theta$ and expression $\beta$ vary, while the latent code and the normal map remains the same. we use a face recognition network's output features to measure whether the two output images correspond to a same person, or known as the identity loss $\mathcal{L}_{\rm id}$,
\begin{align}
	&\mathcal{L}_{\rm id}(n(\alpha,{\beta}_1,{\theta}_{1}), n(\alpha,{\beta}_2,{\theta}_{2}), z) = \nonumber \\
	&\|R(G(n(\alpha,{\beta}_1,{\theta}_{1}), z, {{\varepsilon}})) - R(G(n(\alpha,{\beta}_2,{\theta}_{2}), z, {\varepsilon}))\|_2,
\end{align}
where $R$ is the pre-trained and fixed face recognition network, ${\theta}_{1}$ and ${\theta}_{2}$ denote the different poses fed into the renderer. The human shape $\alpha$ and the latent $z$ remain the same so that we can obtain a same person with the similar facial textures from different view points.

The facial landmark consistency loss $\mathcal{L}_{\rm ldmk}$, together with the identity loss $\mathcal{L}_{\rm id}$, form the contrastive loss, which is designed to further disentangle the input normal map and the latent code $z$. 

Furthermore, to enhance the quality of the rendered image, we also introduced the adversarial loss $\mathcal{L}_{\rm adv}$.
The subsequent loss function $\mathcal{L}_{\rm GAR}$ for the training of the GAR is a weighted sum of the aforementioned losses
\begin{align}
	\mathcal{L}_{\rm GAR} = \lambda_{\rm n}\mathcal{L}_{\rm n} + \lambda_{\rm ldmk}\mathcal{L}_{\rm ldmk} + \lambda_{\rm id}\mathcal{L}_{\rm id} + \lambda_{\rm adv}\mathcal{L}_{\rm adv} 
\end{align}
\noindent
where the $\lambda$s are weights of the corresponding losses.

\subsection{Face Geometry Reconstruction with Generative Adversarial Renderer}

In this section, we introduce our optimization-based framework for face geometry reconstruction, aided by the proposed neural renderer.

\noindent
\textbf{Optimization-based face geometry reconstruction.}
After the neural renderer is trained, it can replace the differentiable renderer in the optimization-based face geometry reconstruction pipeline (indicated by red lines in Fig.~\ref{fig:stylegan}(a)).
Given a test face image $I_{\rm t}$ in the wild, our goal is to reconstruct the face geometry via optimizing the 3DMM parameters $\alpha$, $\beta$ and $\theta$, which can be used to generate the normal map.
The input latent code $z$ as well as the noise ${\varepsilon}$ that encodes other factors of the face image is also optimized.

We first initialize these parameters, and render them with the trained and fixed neural renderer to obtain the rendered image, which is then used to calculate the loss with $I_{\rm t}$. The face geometry reconstruction loss $\mathcal{L}_f$ is defined as
\begin{align}
	\underset{\alpha, \beta, \,{\theta}, z}{\rm minimize}& \,\, \mathcal{L}_f(\alpha, \beta, {\theta}, z, {\varepsilon}) = \|G(\tilde{n}(\alpha, \beta, {\theta}), z, {\varepsilon}) - I_{\rm t}\|_2^2 \nonumber \\
		&+ \sum_i\|F_i(G(\tilde{n}(\alpha, \beta, {\theta}), z)) - F_i(I_{\rm t})\|_2^2 \nonumber \\
		&+ \lambda_{n}||{\varepsilon}||_2^2,
\end{align}
where $G$ represents the fixed generative adversarial renderer, $\tilde{n}$ is the normal map calculated from geometry coefficients ($\alpha,\beta,\theta$), $F_i$ is the $i$th layer feature map by an ImageNet-pretrained VGG network to model the perceptual loss, and $\lambda_n$ weights the regularization term on the random noise.
By minimizing the above face geometry reconstruction loss $\mathcal{L}_f$, we can obtain optimized geometry parameters $\alpha$, $\beta$ and $\theta$.

\begin{table}
\footnotesize
\begin{center}
\begin{tabular}{c|c|c|c}
	\hline
	Method & Condition & CelebA & FFHQ \\
	\hline
	Progressive GAN \cite{karras2017progressive}  & $\times$ & 7.79 & 8.04 \\
	StyleGAN \cite{karras2019style} & $\times$ & 5.17 & 4.40 \\
	Ours & $\checkmark$ & 5.48 & 5.09 \\
	\hline
\end{tabular}

\end{center}
\caption{FID scores of state-of-the-art face generation methods. Even though our GAR is trained with more conditioning constraints and the controllability of the network is promoted, our output images are still competitive to those of the unconditional StyleGAN in terms of image quality and diversity.}
\label{tab:fid}
\vspace{-0.4cm}
\end{table}

\noindent
\textbf{Initialization with Renderer Inverting.}
Although the optimization with random initialization can produce plausible face geometry, we noticed that the gradient-based optimization is likely to get stuck at the local minima of the cost function.
Inspired by \cite{bau2019seeing}, we design a renderer inverting network $V$ to predict a good initial point for the gradient-based optimization of the latent code $z$ to tackle this problem (indicated by green lines in the Fig.~\ref{fig:stylegan}(a)).

The renderer inverting network $V$ and the generative adversarial renderer $G$ are trained in a coupled way, where the output of the neural renderer (the generated face image $I_{\rm out}$) is input into $V$ to convert the image back to a latent code $\hat{z}$. Ideally, the reconstructed latent code $\hat{z}$ should be close to the input latent code $z$.

We design the structure of the inverting network $V$ symmetric to the GAR $G$, which is more theoretically interpretable, with Conv Layers converted to Deconv Layers, and the statistical mean and variance of feature maps are used to estimate the latent code $z$ with an MLP that has same depth and channels for each layers as the style transfer MLP.
The resulting feature maps of the inverting network should be of the same spatial size as those in the corresponding layer of the neural renderer.
The reconstructed latent code $\hat{z}$ is estimated based on the concatenation of each layers' statistic means and standard variances followed by an MLP. The loss function for training the renderer inverting network is therefore
\begin{align}
	\mathcal{L}_z(R) = &\|{\rm MLP}([\mu(R_i(I_{\rm out})); \sigma(R_i(I_{\rm out}))]) - z\|_2^2 \nonumber \\
		&+ \sum_{i} \|G_i( n, z, {\theta}) - R_i(I_{\rm out})\|,
\end{align}
where $I_{\rm out} = G( n, z, {\rm noise})$ is the generated face image from the neural renderer, $R_i$ and $G_i$ denotes the feature map from the $i$th layer of $R$ and $G$ respectively, $\mu, \sigma$ is the mean and variance of the feature map in $R$.

\noindent
\textbf{Initialization with 3DMM Solving.}
To obtain good initial 3DMM face parameters, we adopt the traditional 3DMM fitting algorithm \cite{blanz1999morphable} based on 2D facial landmarks. The loss function is accordingly defined as
\vspace{-0.2cm}
\begin{equation}
\begin{split}
	\mathcal{L}_{3d}
		= \sum_{i}(kRv'_i + t - l_{i})+\left|\frac{s}{\sigma_{s}}\right| + \left|\frac{e}{\sigma_{e}}\right|,
\end{split}
\end{equation}
\vspace{-0.3cm}
	
where $v'_{i} = v_{i} + W^{s}_{i} s + W^{e}_{i} e$ is the location of the vertex. $\alpha, \beta$ are the 3DMM parameters of shape and expressions, $W^s_i, W^e_i$ represent the linear bases of shape and expression in 3DMM model, $k, R, t$ are the pose parameters, and $v_i$ is the $i$-th vertex mean position.


The renderer inverting network, together with the 3DMM parameter pre-solving, provides a good initialization for the optimization. The optimization of the 3D face shape can then be performed by minimizing the photo-metric loss between the rendered image and the input image.


\noindent
\textbf{\label{face_editing}Face editing with the Neural Renderer.} The conventional face reconstruction method can be used for face image editing via modifying the recovered 3D parameters and rendering the modified face geometry. However, the editing's rendered images are generally not realistic, since the reconstruction is not accurate and the rendered images are from the conventional graphics-based renderer.

The proposed face geometry reconstruction method, together with the Generative Adversarial Renderer, provides an effective approach for face editing.
Specifically, given an source image $I_{s}$, the corresponding 3DMM geometry parameters $(\alpha_{s}$, $\beta_{s}$, $\theta_{s})$ can be recovered by our optimization-based framework, as well as the latent code $z_s$ and $\varepsilon_s$.
All or a portion of the 3DMM parameters can be chosen for editing.
By rendering the edited parameters with the Generative Adversarial Renderer, we could obtain the corresponding edited face image $I_t$ with realistic details.
Even though the general idea is similar, the edited faces are much more appealing, owing to the proposed novel renderer.
\section{Experiments}
\begin{figure}[t]
\begin{center}
\includegraphics[width=1.0\linewidth]{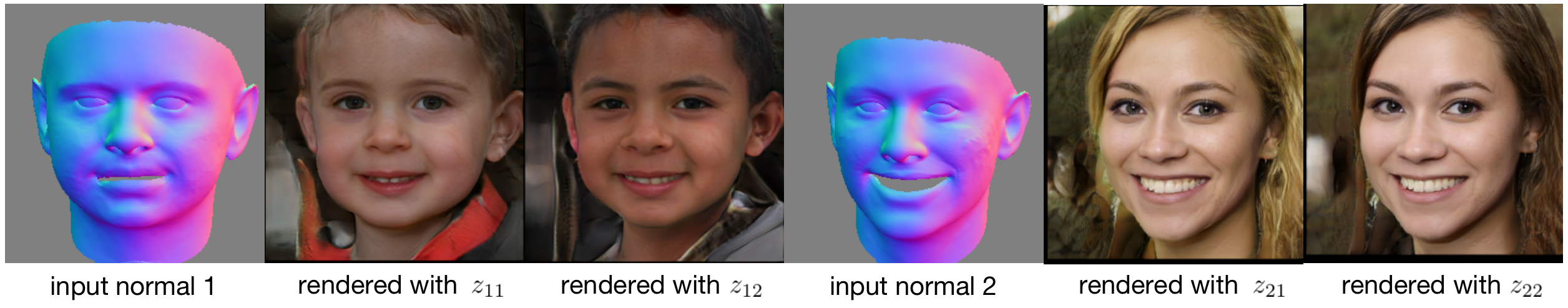}
\end{center}
\caption{Generated Images with same normal and different latent codes. Our GAR renders the input normal map with diverse facial textures encoded by different latent codes.}
\label{fig:render}
\vspace{-5pt}
\end{figure}
\subsection{Dataset}

\begin{figure}[t]
\begin{tabular}{m{0.5cm}<{\centering}|m{.9cm}<{\centering} m{.9cm}<{\centering} m{.9cm}<{\centering} m{.9cm}<{\centering} m{.9cm}<{\centering}}	
\scriptsize
    Inputs
	&\includegraphics[width=1.5\linewidth]{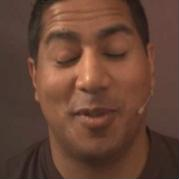}
	&\includegraphics[width=1.5\linewidth]{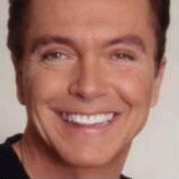}
	&\includegraphics[width=1.5\linewidth]{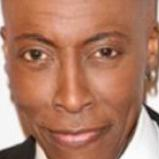}
	&\includegraphics[width=1.5\linewidth]{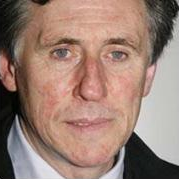}
	&\includegraphics[width=1.5\linewidth]{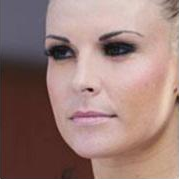} \\
	\hline
	\footnotesize
    \cite{tewari2018self}
	&\includegraphics[width=1.5\linewidth]{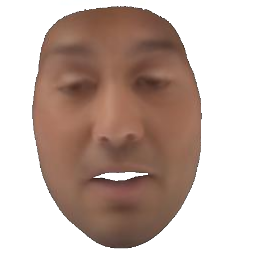}
	&\includegraphics[width=1.5\linewidth]{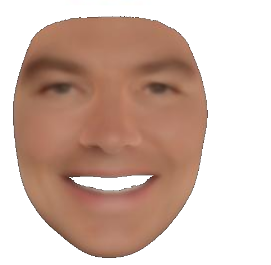}
	&\includegraphics[width=1.5\linewidth]{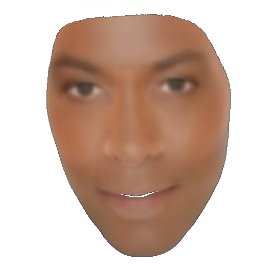}
	&\includegraphics[width=1.5\linewidth]{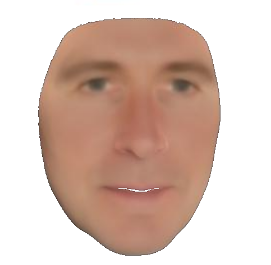}
	&\includegraphics[width=1.5\linewidth]{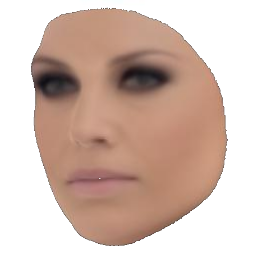} \\
	\footnotesize
    \cite{tuan2017regressing}
	&\includegraphics[width=1.5\linewidth]{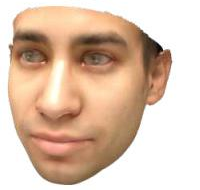}
	&\includegraphics[width=1.5\linewidth]{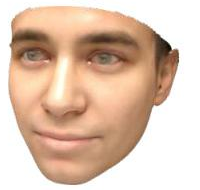}
	&\includegraphics[width=1.5\linewidth]{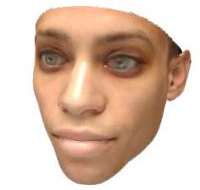}
	&\includegraphics[width=1.5\linewidth]{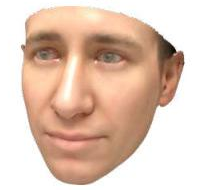}
	&\includegraphics[width=1.5\linewidth]{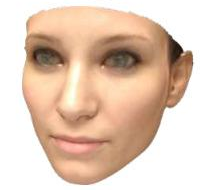} \\
	\footnotesize
    \cite{gecer2019ganfit}
	&\includegraphics[width=1.5\linewidth]{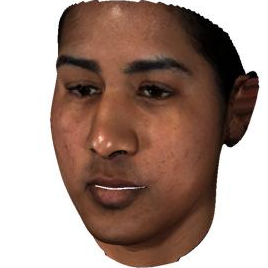}
	&\includegraphics[width=1.5\linewidth]{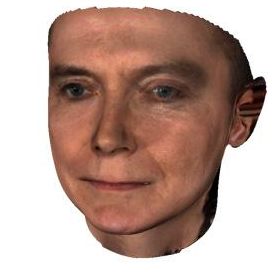}
	&\includegraphics[width=1.5\linewidth]{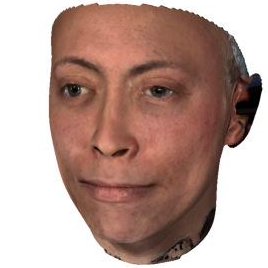}
	&\includegraphics[width=1.5\linewidth]{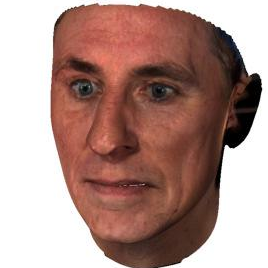}
	&\includegraphics[width=1.5\linewidth]{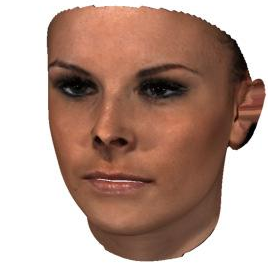} \\
	\footnotesize
	Ours
	&\includegraphics[width=1.5\linewidth]{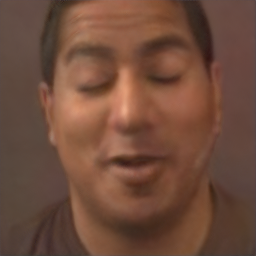}
	&\includegraphics[width=1.5\linewidth]{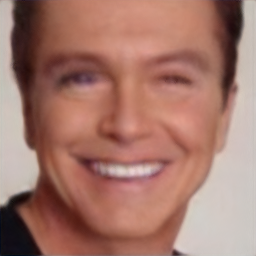}
	&\includegraphics[width=1.5\linewidth]{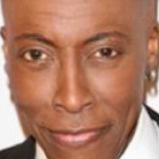}
	&\includegraphics[width=1.5\linewidth]{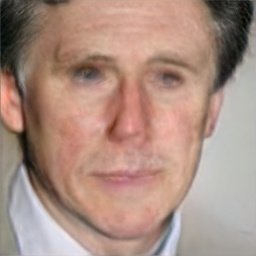}
	&\includegraphics[width=1.5\linewidth]{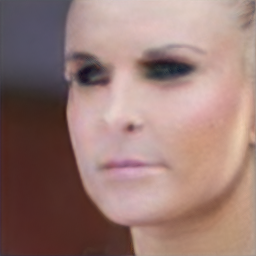} \\
	\hline
    \footnotesize
    \cite{tewari2018self}
	&\includegraphics[width=1.5\linewidth]{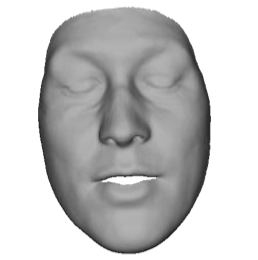}
	&\includegraphics[width=1.5\linewidth]{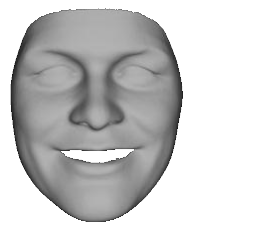}
	&\includegraphics[width=1.5\linewidth]{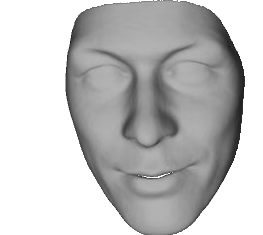}
	&\includegraphics[width=1.5\linewidth]{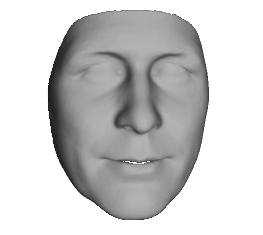}
	&\includegraphics[width=1.5\linewidth]{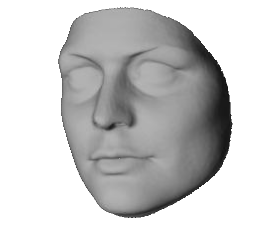} \\
    \footnotesize
    \cite{tran2018nonlinear}
	&\includegraphics[width=1.5\linewidth]{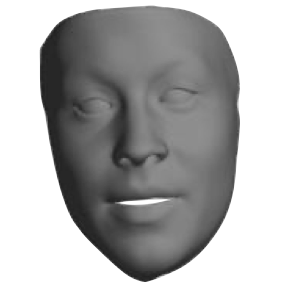}
	&\includegraphics[width=1.5\linewidth]{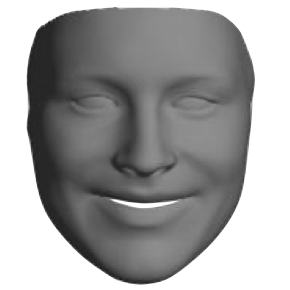}
	&\includegraphics[width=1.5\linewidth]{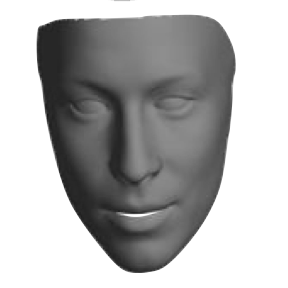}
	&\includegraphics[width=1.5\linewidth]{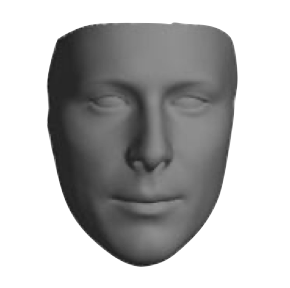}
	&\includegraphics[width=1.5\linewidth]{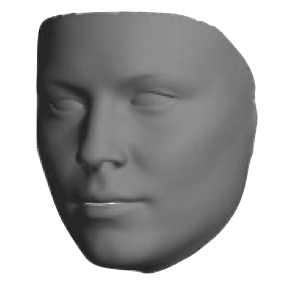} \\
    \footnotesize
    \cite{gecer2019ganfit}
	&\includegraphics[width=1.5\linewidth]{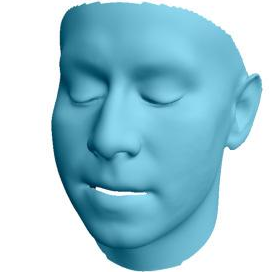}
	&\includegraphics[width=1.5\linewidth]{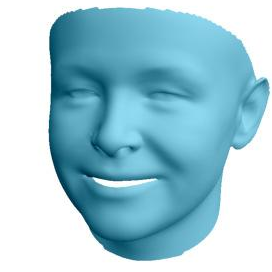}
	&\includegraphics[width=1.5\linewidth]{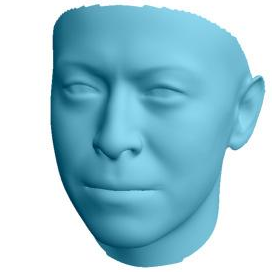}
	&\includegraphics[width=1.5\linewidth]{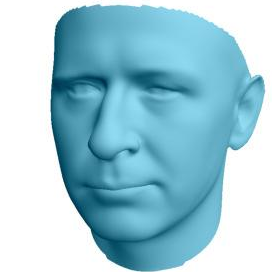}
	&\includegraphics[width=1.5\linewidth]{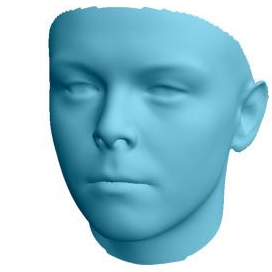} \\
	\footnotesize
	Ours
	&\includegraphics[width=1.5\linewidth]{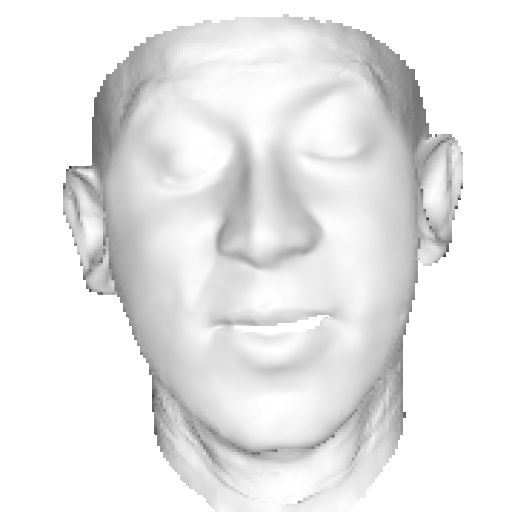}
	&\includegraphics[width=1.5\linewidth]{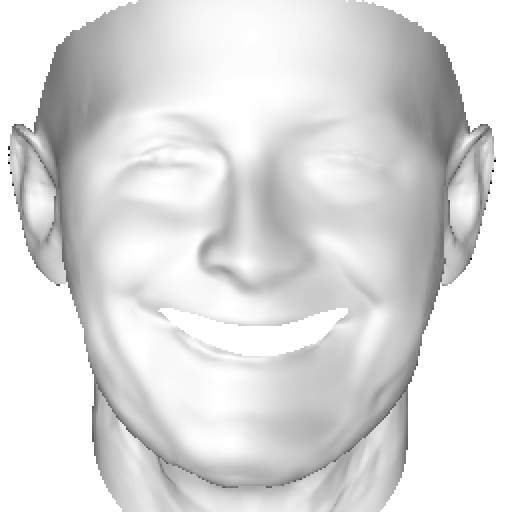}
	&\includegraphics[width=1.5\linewidth]{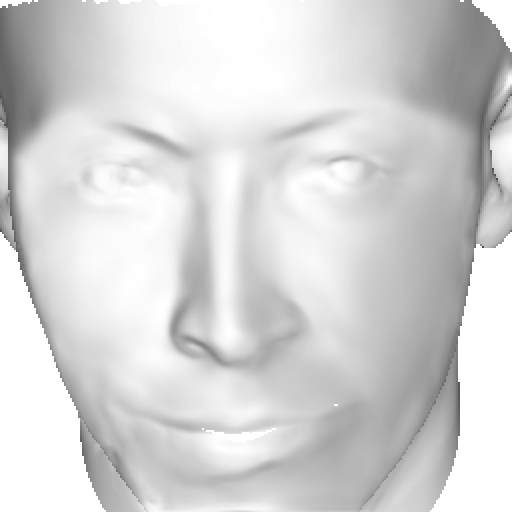}
	&\includegraphics[width=1.5\linewidth]{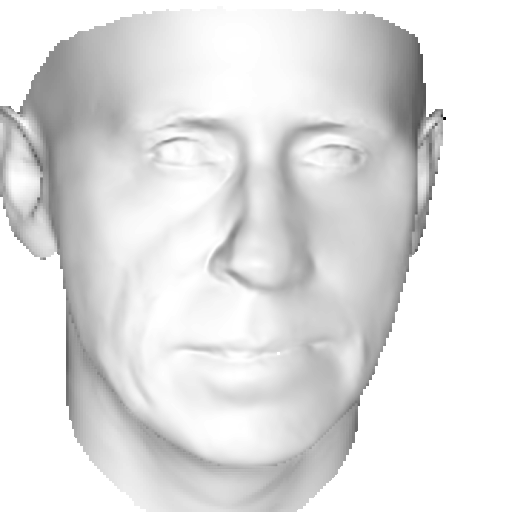}
	&\includegraphics[width=1.5\linewidth]{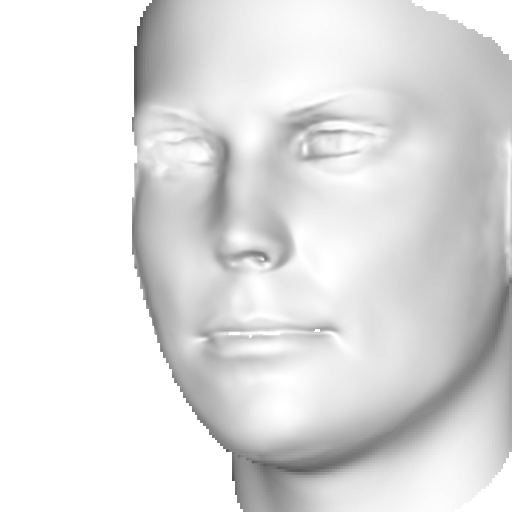} \\
\end{tabular}
\caption{Reconstructed geometry (bottom) and rendered images (top) compared to previous methods. Our results outperforms others by a large margin, in both the geometry accuracy and the similarity of the rendered images.}
\label{fig:reconstruct}
\vspace{-5pt}
\end{figure}

Our algorithm is trained in a self-supervised manner, and requires no image annotated with 3DMM parameters.
To train and test our proposed algorithm's performance, the following datasets are adopted.

\noindent
\textbf{Flickr-Faces-High-Quality (FFHQ)}~\cite{karras2019style} is a dataset of aligned faces in resolution of $1024\times 1024$ with labeled facial landmarks. The dataset covers larger variations of face orientations, backgrounds than other high-resolution datasets. 
We adopt this dataset to train our generative adversarial renderer.

\noindent
\textbf{CelebFaces Attribute (CelebA)}~\cite{liu2018large} is a dataset of celebrities with more variations, which contains challenging cases for reconstruction. This dataset is used for the self-supervised training of the normal estimation network \cite{sengupta2018sfsnet}, and a subset of this dataset is used to evaluate the quality of the rendering results.

\noindent
\textbf{MoFA Dataset (MoFA)}~\cite{tewari2017mofa} is a combination of four datasets~\cite{cao2014facewarehouse,huang2008labeled,liu2015deep,shen2015first}, including unlimited faces in widely-ranged circumstances. We follow \cite{tewari2017mofa} and \cite{gecer2019ganfit} to evaluate our method qualitatively on its testset.

\noindent
\textbf{Florence 3D Faces (Florence)}~\cite{bagdanov2011florence} includes several scanned 3D face meshes of $53$ subjects. The three videos about these subjects are provided, which are taken from outdoor, indoor, and cooperative environments.

\subsection{Implementation Details.}

For all of our experiments, a given face image is aligned to our fixed template using 68 landmark locations~\cite{wu2018look, qian2019aggregation, sun2019fab} detected by an hourglass 2D landmark detection~\cite{newell2016stacked}. For the normal map estimation, we adopt SFSNet~\cite{sengupta2018sfsnet}.

During the GAR training process, we optimize parameters using Adam solver with a 0.01 learning rate.
We set our balancing factors as $\lambda_{\rm n}=2.0$, $\lambda_{\rm ldmk}=2.0$, $\lambda_{\rm id}=1.0$, $\lambda_{\rm adv}=1.0$.

For the evaluation of the Florence dataset, we uniformly sample $5$ frames of each video, and calculate the average of the vertex coordinates for evaluation following~\cite{gecer2019ganfit}.
The evaluation metric for face reconstruction is the point-to-plane error of each vertex of reconstructed 3DMM meshes to the ground-truth scanned meshes.

\subsection{Evaluation on Face Image Generation}

The examples of generated images by our approach can be seen in Figure \ref{fig:render}, which shows that the proposed Generative Adversarial Renderer can generate face images with much higher visual quality than conventional graphics-based renderers, where random hairstyle, glasses, and other attributes are well generated.
To quantitatively analyze the image quality of the generated images, we randomly generated $50,000$ images with Progressive GAN~\cite{karras2017progressive}, StyleGAN~\cite{karras2019style} and our GAR, and calculate the Frechet Inception Distance (FID) \cite{shmelkov2018good} between the generated images and the real image datasets. The results are presented in Table~\ref{tab:fid}.
It demonstrates that even though our GAR is trained with more conditioning constraints and the controllability of the network is promoted, our output images are still competitive to those of the unconditional StyleGAN in terms of image quality and diversity.

For the input conditioning normal maps in Fig.~\ref{fig:render}, we can see that the face geometry is well maintained during the rendering. This indicates that our GAR follows the important role of a renderer, faithfully converting an input normal map and a latent code to a corresponding face image.



\begin{table}
\scriptsize
\begin{center}
\begin{tabular}{c|cc|cc|cc}
	\hline
	\multirow{2}{*}{Method} &
	\multicolumn{2}{c|}{Cooperative} &
	\multicolumn{2}{c|}{Indoor} &
	\multicolumn{2}{c}{Outdoor} \\
	& Mean & Std & Mean & Std & Mean & Std \\
	\hline
	Tran et al.\cite{tuan2017regressing}   & 1.93 & 0.27 & 2.02 & 0.25 & 1.86 & 0.23 \\
	Booth et al. \cite{booth20173d} & 1.82 & 0.29 & 1.85 & 0.22 & 1.63 & 0.16 \\
	Genova et al. \cite{genova2018unsupervised}& 1.50 & 0.13 & 1.50 & 0.11 & 1.48 & 0.11 \\
	GANFIT \cite{gecer2019ganfit}& 0.95 & \textbf{0.11} & 0.94 & 0.11 & 0.94 & 0.11 \\
	\hline
	Ours w/o norm-cycle & 1.20 & 0.31 & 1.10 & 0.33 & 1.40 & 0.53 \\
	Ours w/o initial    & 3.20 & 2.10 & 3.21 & 1.97 & 2.98 & 1.43 \\
	Ours                & \textbf{0.94} & 0.12 & \textbf{0.92} & \textbf{0.11} & \textbf{0.90} & \textbf{0.08} \\
	\hline
\end{tabular}
\end{center}
\caption{Reconstruction errors of meshes in terms of point-to-plane distance on Florence dataset.}
\label{tab:error}
\end{table}

\subsection{Evaluation on Face Geometry Reconstruction}

\noindent
\textbf{Qualitative Comparison.}
We qualitatively compare our reconstruction algorithm with several state-of-the-art methods on the MoFA-Test dataset (see Figure \ref{fig:reconstruct}).
Rows 6 to 9 are reconstructed face meshes. Our mesh results are apparently more accurate in terms of both shape and expression, with more high-fidelity details.
Row 2 to Row 5 are rendered images. 
Our rendered images are very close to the input images, since we significantly narrow the gap between the rendered and the realistic images.

\noindent
\textbf{User Study.}
We also conducted a user study to ask people to vote for a reconstruction result most similar to the input image. The results show that 59.1\% users believe our result is the most consistent with the target image, while the second best~\cite{tewari2018self} has only 28.2\%, which verifies the superiority of the expression ability of our algorithm.

\begin{figure}[t]
\begin{center}
\includegraphics[width=1\linewidth]{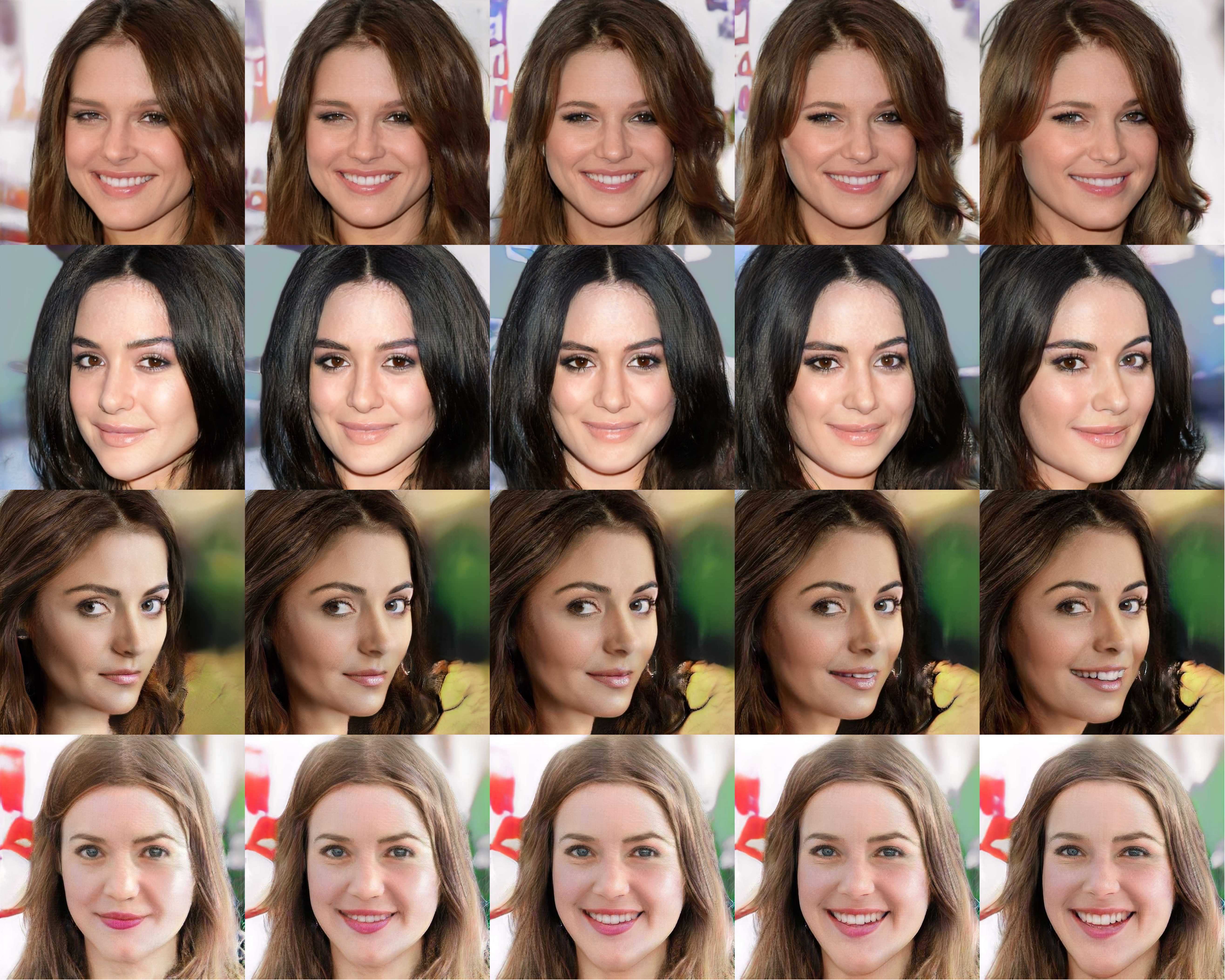}
\end{center}
\caption{Face Editing Effects. By editing the 3DMM parameters, the rendered image would present corresponding attributes. In Row 1 and 2, the pose is set to turn from left to right, so the faces in the rendered images gradually changes while the identity and the expression maintain unchanged. In Row3 and 4, we present the resulting images of editing face expressions.}
\label{fig:edit}
\vspace{-5pt}
\end{figure}

\noindent
\textbf{Quantitative Comparison.}
To quantitatively evaluate our reconstruction algorithm's performance, we use scanned human faces to test the accuracy. We use 5 frames from each video in the Florence Dataset \cite{bagdanov2011florence} and compare the result to the ground-truth scanned meshes. The results are shown in Table \ref{tab:error}.
Since our mesh is calculated in the camera space, we perform an ICP (iterative-closest-point) algorithm to align the output mesh by our method to the scanned ground-truth. The errors are calculated as the point-to-plane distance for each vertex on our reconstructed meshes.
Our method has a better result in terms of the average error.

\subsection{Ablation Study}

In this section, we present the results of an ablation study on investigating different components of our proposed face reconstruction framework.


\noindent
\textbf{Effect of Normal Consistency Loss.}
As shown in Line 5 and Line 7 of the Table~\ref{tab:error}, when training GAR without the Normal Consistency Loss (``Ours w/o norm-cycle''), the training cannot guarantee the results of the generator to well condition on the input normal maps. This indicates that the proposed Normal Consistency Loss is significant in promoting the controllability of the input normal maps.

\noindent
\textbf{Effect of Renderer Inverting Initialization.}
As shown in Line 6 and Line 7 of the Table~\ref{tab:error}, when the latent code is not initialized with the renderer inverting (``Ours w/o initial''), the reconstruction error shows a severe increase and might not converge for specific faces.

\noindent
\textbf{Choices of the Conditions.}
Previous works tend to employ 3DMM parameters or depth as the conditioning inputs. However, we found that the face normal map is a more effective form of condition for our GAR. With the face normal maps as inputs, the loss is minimized more stably and faster, and the network could converge to a better optimum.

\noindent
\textbf{Effect of the Normal Injection Module (NIM).}
Comparing with the simple concatenation of the normal map into the feature maps, the proposed NIM is effective in further minimizing the loss value, which demonstrates the effectiveness of the proposed NIM.

\subsection{Qualitative Evaluation on Face Image Editing}

As mentioned in Section~\ref{face_editing}, our method is capable of face editing.
As shown in Figure~\ref{fig:edit}, by editing the 3DMM parameters, the rendered image would present corresponding attributes. In the first row, the pose is set to turn from left to right, so the faces in the rendered images gradually change while the identity and the expression maintain unchanged. In the second and third rows, we present the resulting images of editing facial expressions.
For results please refer to the supplementary materials.

\section{Conclusion and Future Work}

In this paper, we propose a Generative Adversarial Renderer (GAR) that takes a normal map and a latent code and outputs a rendered face image. Based on GAR, we also propose an optimization-based face geometry reconstruction method, as well as an initialization method by Renderer Inverting.


The main idea of this paper may be naturally extended to arbitrary scenarios. For instance, we may train a generative adversarial renderer for bedrooms, which takes normal maps of a bed and then renders corresponding images. The renderer might also be used to reconstruct their geometry.
\\
\par\noindent\textbf{Acknowledgement} 
This work is supported in part by the General Research Fund through the Research Grants Council of Hong Kong under Grants (Nos. 14208417 and 14207319), in part by CUHK Strategic Fund.

{\small
\bibliographystyle{ieee_fullname}
\bibliography{egbib}
}

\end{document}